\documentclass[runningheads]{llncs}
\usepackage[T1]{fontenc}
\usepackage{graphicx}
\usepackage{fancyhdr}
\fancypagestyle{empty}{%
  \fancyhf{} 
  \fancyfoot[C]{\textit{This paper has been submitted to the upcoming 18th International Conference on Advances in Social Networks Analysis and Mining (ASONAM 2026)}}
  
}

\begin{document}
\title{LLM-based Detection of Manipulative Political Narratives}
%
%

\author{Sinclair Schneider\inst{1}\orcidID{0000-0002-6332-7014} \and
Florian Steuber\inst{1}\orcidID{0000-0003-1782-7445} \and
Gabi Dreo Rodosek\inst{1}\orcidID{0000-0002-8702-8553}}

\authorrunning{S. Schneider et al.}
%
\institute{University of the Bundeswehr Munich\\
Werner-Heisenberg-Weg 39, 85579 Neubiberg, Germany
\email{sinclair.schneider@unibw.de}, \email{florian.steuber@unibw.de}, \email{gabi.dreo@unibw.de}\\
}
\maketitle              
\begin{abstract}
We present a new computational framework for detecting and structuring manipulative political narratives. A task that became more important due to the shift of political discussions to social media. One of the primary challenges thereby is differentiating between manipulative political narratives and legitimate critiques. Some posts may also reframe actual events within a manipulative context.

To achieve good clustering results, we filter manipulative posts beforehand using a detailed few-shot prompt that combines documented campaign narratives with legitimate criticisms to differentiate them. This prompt enables a reasoning model to assign labels, retaining only manipulative narrative posts for further processing.

The remaining posts are subsequently embedded and dimensionality-reduced using UMAP, before HDBSCAN is applied to uncover narrative groups. A key advantage of this unsupervised approach is its independence from a predefined list of target categories, enabling it to uncover new narrative clusters.

Finally, a reasoning model is employed to uncover the narrative behind each cluster. 
This approach, applied to over 1.2 million social media posts, effectively identified 41 distinct manipulative narrative clusters by integrating prompt-based filtering with unsupervised clustering.

\end{abstract}
\section{Introduction}

\begin{quote}
Strategic narratives are a means for political actors to construct a shared meaning of the past, present, and future of international politics to shape the behavior of domestic and international actors \cite[p.~3]{miskimmon2013strategic}. 
\end{quote}
For instance, during the Second World War, the British disinformation radio station ``Gustav Siegfried Eins'' successfully deployed fabricated narratives of elite corruption to drive a wedge between frontline soldiers and their leadership \cite[pp.~64--65]{delmer1962}. By contrasting the honorable sacrifices of the past with the fabricated present reality where party elites are living a luxurious life while soldiers freeze, the broadcaster projects a future of pointless deaths, leading to weak troop morale. While the basic building blocks of manipulative content, such as moral inversion, blame-shifting, and fabricated elite betrayal, remained remarkably consistent, the dissemination channels have changed. Modern Foreign Information Manipulation and Interference (FIMI) campaigns have shifted from centralized broadcasting to algorithmic amplification on social media platforms to inject manipulative content directly into adversaries' domestic political discourse \cite{1stEEASReport,wardle2017information}.

Modern state-aligned campaigns employ advanced techniques for the injection of manipulative content. For example, campaigns such as ``Doppelgänger'' rely on cloning legitimate news outlets to deliver disinformation \cite{alaphilippe_doppelganger_2022}, while ``Storm-1516'' employs narrative laundering and the production of synthetic scandals through forged evidence and staged videos \cite{NarrativeLaundering2023,BugattiFirstLady2024}.

Consequently, the automated detection of FIMI presents a critical challenge for modern computational social science. Effective manipulation rarely relies solely on falsehoods (disinformation) but often evolves from malicious reframing of factual events (malinformation) to fit a specific agenda. Therefore, rather than strictly fact-checking claims for truthfulness, this paper focuses on detecting the overarching manipulative intent and rhetorical motifs that characterize these strategic narratives, regardless of their strict factual veracity. The key task, then, is separating these coordinated, manipulative storylines from legitimate yet highly controversial political critique. Traditional classification approaches and standard topic modeling techniques often fail to capture the underlying manipulative intent by overlooking the rhetorical nuances that characterize these campaigns.

In response to these limitations, this paper addresses the research question: \textit{How can politically manipulative strategic narratives be identified and structured within an unfiltered, large-scale dataset of social media posts?}

We propose a Large Language Model (LLM) driven data-processing pipeline for detecting and clustering FIMI narratives. To ensure precise detection, we use FIMI characteristics and a few-shot set of examples to guide a reasoning model in identifying the nuances that distinguish legitimate political critique from manipulative content. After mapping the identified posts into an embedding space structured around their underlying motives, density-based clustering is applied to uncover new narrative groups without relying on a predefined list.

This work presents three contributions:

\textbf{Prompt-Based Reasoning:} Going beyond traditional BERT-based classification, we introduce a prompt-based reasoning approach. By guiding the model with explicit FIMI characteristics and few-shot examples, this method successfully isolates strategic manipulative content from legitimate political critique based on rhetorical nuances.

\textbf{Intent-Driven Embedding:} To shift the focus of the original BERTopic \cite{bertopic2022} pipeline from topics to narratives, the embedding model is explicitly configured to map posts based on their manipulative intent. This adjustment ensures that related storylines are close together in the embedding space.

\textbf{Strategic narrative extraction: } We replace the standard topic extraction mechanism with a specialized prompt designed to capture FIMI-related strategic narratives. By instructing the model to include the core claim, the targeted adversary, and the manipulative angle, we extract complete storylines rather than simplistic topical keywords.

\section{Related Work and Fundamentals}
Research on political disinformation and malinformation narratives primarily concentrates on two key areas. The first area involves creating datasets that provide a foundation for further exploration. The second area focuses on applying topic modeling techniques to established corpora of this manipulative content. 

These datasets include sources such as Reliable Recent News (rrn.world) and WarOnFakes (waronfakes.com), as well as linguistic analyses that compare the two and use unsupervised topic clustering \cite{heppell2023}. Furthermore, several dataset publications focus on collecting human-annotated posts on specific topics, such as elections in the United Kingdom \cite{haouari2025uk}.

Closest to our approach is DiNaM (Disinformation Narrative Mining with Large Language Models)\cite{sosnowski2025dinam}, which similarly implements an LLM-assisted pipeline with a final clustering step. However, DiNaM operates on fact-check articles, whereas our approach targets unfiltered social media posts, where manipulative narratives must first be separated from legitimate political critique and unrelated content. 
These studies share the common feature of using predefined corpora of manipulative content that may also be related to Covid-19 \cite{schaefer2024bertopic} or the spread of known Russian state media narratives on Reddit \cite{hanleyHappenstanceUtilizingSemantic2023}.
Although the scope of this paper is limited to text processing, there are modeling approaches that combine text and images using BERTopic with CLIP \cite{steffen2025memes}.


\subsection{Information Disorders, FIMI and Strategic Narrative}
To clarify the foundations of our methodology section, we briefly introduce the fundamental terms and explain their interactions.
\subsubsection{Information Disorders} are separated by Wardle et al. in the following three categories: \cite[p. 20]
{wardle2017information}

\begin{itemize}
    \item \textbf{Dis-information.} Information that is false and deliberately created to harm a person, social group, organization or country.
    \item \textbf{Mis-information.} Information that is false, but not created with the intention of causing harm.
    \item \textbf{Mal-information.} Information that is based on reality, used to inflict harm on a person, organization or country.
\end{itemize}

\subsubsection{FIMI} (Foreign Information Manipulation and Interference) is defined by the European External Action Service (EEAS) as ``a mostly non-illegal pattern of behavior that threatens or has the potential to negatively impact values, procedures and political processes.'' \cite{1stEEASReport} 

\subsubsection{Strategic Narratives} are according to Miskimmon et al. ``a means for political actors to construct a shared meaning of the past, present, and future of international politics to shape the behavior of domestic and international actors'' \cite{miskimmon2013strategic}.

Consider the disinformation narrative that accuses the Ukrainian government of being engaged in trafficking children to the West\cite{DisinfoUkrainianChildren}. 

\begin{itemize}
    \item \textbf{Past:} Ukraine has a history of corruption and inhumanity.
    \item \textbf{Present:} Children are suffering, with implied Western complicity.
    \item \textbf{Future:} Ukraine is deemed unworthy of support, justifying a brutal war.
\end{itemize}

A narrative is more than just a topic, it is a story that shapes our understanding of the world. This paper focuses on strategic narratives designed to influence the actions of domestic and international actors.

\subsubsection{The overall interaction} 
\begin{figure}
\includegraphics[width=\textwidth]{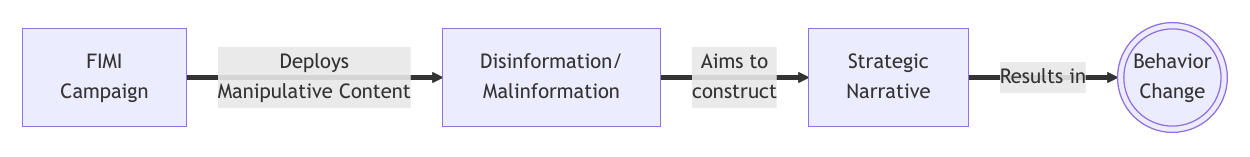}
\caption{The way from a FIMI campaign to a behavior change at the audience} \label{figFIMI}
\end{figure}

of a FIMI campaign ranges from the deployment of the manipulative content to the intended behavioral change of the audience, as shown in Figure \ref{figFIMI}.

During a FIMI campaign, manipulative content, such as disinformation, is disseminated through channels such as Telegram, X, and Reddit to shape audience behavior. For example, a campaign might falsely claim that Ukraine is trafficking children to the West, reinforcing negative perceptions of Ukrainian corruption and depicting children as victims. This strategy seeks to shift audience sentiment, increasing opposition to future support for Ukraine.

\subsection{Real-World Influence Operations}

The execution of these strategic narratives can be best understood by analyzing the tactics employed in recent large-scale FIMI operations.

\subsubsection{Doppelgänger} refers to a Russian disinformation campaign using lookalike news outlets. The European Union’s Disinformation Lab reports that the Russian Social Design Agency (SDA) and Structura National Technologies have created at least 17 cloned sites, such as Bild and The Guardian, along with a fake NATO site at nato[.]ws and a pro-Russian outlet at RNN[.]media, which promotes ``fact-checked'' content \cite{alaphilippe_doppelganger_2022}. 

FIMI delivery mechanisms range from website cloning to fabricated whistleblowers, with campaigns relying on common rhetorical motifs to turn political ambiguity into malicious storylines. Table \ref{tab:fimi_campaigns} outlines real-world FIMI campaigns and their motifs, helping establish criteria for the few-shot queries in our detection pipeline.

\begin{table}[!htbp]
\centering
\caption{Overview of Real-World FIMI Campaigns and Operational Motifs.}
\vspace{2mm}
\begin{tabular}{p{2.5cm} p{9.5cm}}
\hline
\textbf{Campaign} & \textbf{Rhetorical Motifs and Strategies} \\
\hline
\raggedright \textbf{Doppelgänger} & Cloning of news outlets to inject misleading content. \newline
- \textit{Economic pain:} Sanctions destroy domestic economy \cite{meta2022cib,vignium2024rnn} \newline
- \textit{Identity threat:} Russia victim of ``Russophobia'' \cite{DisinfoUkraineNeoNazi,vignium2024rnn} \newline
- \textit{Atrocity propaganda:} Demonizing opponents (Nazis) \cite{spring2022bbc,vignium2024rnn} \newline
- \textit{Refugee scapegoating:} Blaming migrants (destabilization) \cite{syedHowOnlineMisinformation2024} \\
\hline
\raggedright \textbf{Storm-1516} & Disinformation network specialized in creating synthetic scandals. \newline
- \textit{Leaked invoices:} Character assassination (misappropriation) \cite{BugattiFirstLady2024}\newline
- \textit{Staged videos:} Manufacturing credibility for slandering \cite{NarrativeLaundering2023} \\
\hline
\raggedright \textbf{Voice of Europe} & Information laundering outlet that amplifies unserious claims. \newline
- \textit{Credibility piggybacking:} Exploit parliamentarians authorities \cite{VoiceOfEuropePISM}\\
\hline
\raggedright \textbf{White Propaganda} & Legitimizing authorities by positive narratives \newline
- \textit{Positive legitimation}: Development of security and harmony \cite{usstate2023prc}\\
\hline
\raggedright \textbf{Hyper-local FIMI} & Exploit local conflicts to undermine national institutions \newline
- \textit{Conflict reframing}: E.g. localised campaigns \cite{3rdEEASReport}\\
\hline
\end{tabular}
\label{tab:fimi_campaigns}
\end{table}

\subsection{Use of Gray Literature and Source Selection}
To analyze recent strategic narratives, it's vital to examine active FIMI operations. The rapid evolution of political manipulation tactics in social media often outpaces academic literature, making high-quality gray literature essential for understanding current threats. We selected credible gray literature from public institutions, security agencies, fact-checking initiatives, research organizations, and reliable news outlets. While these sources complement peer-reviewed literature, they primarily document recent FIMI tactics and narratives.

\section{Dataset}
The unfiltered dataset used in this paper comprises 1,255,895 short social media posts collected from X (formerly Twitter), Reddit, and Telegram, with an 80\% German and 20\% English split. X accounts for the largest portion, featuring 829,191 tweets. This dataset was compiled by searching for the names of all politicians in the German Bundestag between January and February 2025, prior to the last federal elections in Germany in February 2025. The majority of tweets focused on the right-wing AfD leader Alice Weidel (26.26\%), followed by the social democrat and former health minister Karl Lauterbach (16.61\%), and the newly elected German chancellor Friedrich Merz (7.64\%). 

Reddit was examined using the names of German political parties, politicians, and popular political channels, resulting in a total of 362,753 posts. The leading sources on Reddit were the left-leaning content creator Staiy (9.47\%), neoliberal (8.68\%), and the German left-wing party ``Die Linke'' (6.84\%). These distributions suggest that, within our collected Reddit sample, left-leaning sources were more prominent than in the X subset.

In contrast, Telegram operates on a group-based system rather than open discussion forums or threads, which requires us to join specific groups. This approach yielded 63,951 messages from 219 Telegram groups. Consequently, we mostly engaged with right-wing conspiracy groups, such as SchubertsLM (8.53\%) and EvaHermanOffiziell (5.99\%) \cite{mueller2022telegram}. As a result, the Telegram portion of the dataset reflects a more selective sampling strategy than the other two platforms.

\begin{figure}
\includegraphics[width=\textwidth]{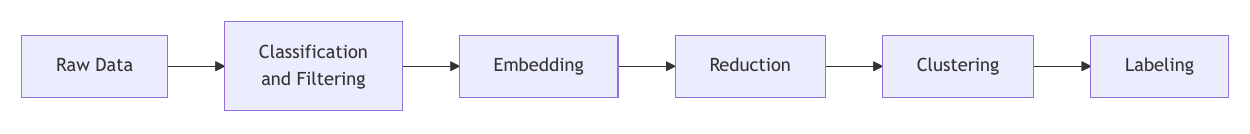}
\caption{Data flow, from raw data to labels} \label{figDataFlow}
\end{figure}

Figure \ref{figDataFlow} provides an overview of the data flow from raw data to narrative labels. All individual steps are described in Section \ref{secMethodology}.

\section{Methodology} \label{secMethodology}
To effectively identify and group manipulative content, we introduce a specialized data-processing pipeline. This approach begins with a filtering step to isolate relevant candidate posts, which are subsequently processed through an adapted BERTopic architecture \cite{bertopic2022} to form cohesive strategic narrative clusters.

\subsection{Prompt-based Filtering}

The goal of prompt-based filtering is to eliminate posts that either provide valid critiques or are unrelated to the topic, while retaining only those that resemble established manipulative campaigns. Since concepts such as blame-shifting, victimhood, and moral inversion apply across various fields, the prompt is broadly applicable. We use an iterative refinement process that combines human expertise and machine optimization. Human experts define strategic narratives and provide examples of relevant campaigns. An LLM (Gemini) then reformulates this knowledge into a structured prompt. This cycle of evaluation and refinement continues until the prompt effectively captures FIMI concepts with a diverse array of few-shot examples.

The prompt is processed using the Qwen3.5-122B-A10B-FP8 model \cite{qwen35} in conjunction with the vLLM \cite{kwon2023pagedattention} inference service, with applied reasoning. We opted for the second-largest model, with 122 billion parameters, because it requires either two Nvidia H200 GPUs or four H100 GPUs to operate. 
This setup provides an ideal trade-off between robust reasoning performance and concurrent throughput on a single multi-GPU node.
Additionally, the mixture-of-experts design of the chosen model means that, despite the 122 billion total parameters, only 10 billion are activated at any given time. This represents a trade-off between the ability to reason with a very complex prompt and the efficient processing of over 1,000,000 posts in a reasonable time. Ultimately, the prompt, as schematically illustrated in Table \ref{tab:prompt_architecture}, is processed, and responses are filtered to exclude invalid outputs. 



\begin{table}[ht]
\centering
\caption{Core building blocks and constraints of the FIMI detection prompt.}
\label{tab:prompt_architecture}
\renewcommand{\arraystretch}{1.3} 
\begin{tabular}{@{} p{0.20\linewidth} p{0.75\linewidth} @{}}
\hline
\raggedright \textbf{Component} & \textbf{Instruction} \\
\hline
\raggedright \textbf{Persona Assignment} & The model is instructed to be an ``AI threat intelligence and FIMI analyst'' to guide its reasoning into geopolitical manipulation. \\
\raggedright \textbf{Analytic Scope} & FIMI was strictly defined as an intentional, coordinated behavior pattern aimed at polarization. The core mechanic is detecting attempts to ``collapse uncertainty'' into a malicious plot. \\
\raggedright \textbf{Known Motifs} & The prompt requires the text to align with at least one documented campaign motif: Doppelgänger, Storm-1516, Voice of Europe, White Propaganda, or Hyper-Local FIMI. \\
\raggedright \textbf{False Positive Constraint} & To prevent overly strict classification, normal government criticism, policy skepticism, and personal economic frustration were explicitly instructed to be evaluated as false. \\
\raggedright \textbf{Output Parameters} & The model was constrained to output exclusively valid JSON (a boolean \texttt{contains\_narrative} classification and a \texttt{reasoning} string). \\
\hline
\end{tabular}
\end{table}

\subsection{Embedding-Generation} \label{sec:embedding}
After filtering the posts, the next step is to map them into the embedding space. While classical sentence transformer models such as all-MiniLM-L6-v2 \cite{reimers2019sbert} are commonly employed, we chose the Qwen3-Embedding-8B \cite{zhang2025qwen} model for two main reasons. First, this model is highly ranked on the Massive Text Embedding Benchmark (MTEB) leaderboard \cite{muennighoff2023mteb}. More importantly, unlike models such as all-MiniLM-L6-v2, Qwen3-Embedding-8B enables us to influence its placement in the embedding space via a specific prompt. This feature is essential because our use case differs from that of a typical topic model. For our instruction prompt, we utilized:

``Identify the strategic narrative, manipulative intent, and underlying disinformation motive in the following text: ''

Although the resulting embeddings feature a high dimensionality of 4096, the initial filtering stage sufficiently reduces the dataset volume to maintain computational efficiency. After generation, the vectors are L2-normalized.

\subsection{Dimensionality Reduction using UMAP}

To conduct subsequent unsupervised clustering, it is essential to reduce dimensionality. We use the Uniform Manifold Approximation and Projection for Dimension Reduction (UMAP) algorithm \cite{mcinnes2018umap} to visualize the clusters in two dimensions and to perform unsupervised clustering in five dimensions. This five-dimensional approach aligns with the standard used in the BERTopic framework. We opt for this limited dimensionality because, even after filtering, retaining only 10\% of the data may still result in a cluster containing around 100,000 posts, making higher-dimensional clustering computationally demanding. Additionally, we maintain the default parameters for minimum distance (set to 0) and the number of approximate nearest neighbors (set to 15), as recommended.

\subsection{Clustering using HDBSCAN}

The choice of clustering algorithm and its hyperparameters is crucial for our analysis. Given our limited knowledge of the resulting clusters and their number, the Hierarchical Density-Based Spatial Clustering of Applications with Noise (HDBSCAN) algorithm \cite{campello2013hdbscan} is the most suitable solution. We will allow the algorithm to ascertain the number of clusters based on the specified hyperparameters. To determine the optimal minimum cluster size, we tested the values 100, 200, 400, 600, 800, and 1000. 
The right minimum cluster size can be determined later by checking if the resulting narratives overlap. 
HDBSCAN also includes the min\_samples parameter, which specifies the minimum number of data points required within a radius of $\epsilon$ for a point to qualify as a core point of a cluster. Since the default value is the minimum cluster size, it is often too high, resulting in only a few clusters, if any. Generally, a higher min\_samples parameter yields more conservative clustering, leading to more points being classified as noise. If the value is set too high, no clusters will be detected, and conversely, if it is set too low, an excessive number of noisy clusters may emerge. 
To focus on highly coherent clusters in the final narrative extraction, we set the min\_samples parameter to 100.

\subsection{Narrative Labeling}
In the final step, it is crucial to establish a narrative for each cluster. Following the standard procedure of the BERTopic framework, we generate a list of keywords using c-TF-IDF (Class-based Term Frequency-Inverse Document Frequency) \cite{bertopic2022} and provide this list to a reasoning model, along with the documents associated with the relevant cluster. Because the limited number of resulting clusters significantly reduces the inference burden compared to the initial filtering stage, we deploy the larger-scale Qwen3.5-397B-A17B-FP8 model \cite{qwen35} for this final extraction. In contrast to conventional topic modeling methods, the prompt used to generate the final narratives for each cluster differs. Similar to the prompt used in the filtering step, we employed a few-shot design to guide the language model toward the desired output, as demonstrated in Table \ref{tab:topicPrompt}.


\begin{table}[ht]
\centering
\caption{Core building blocks of the narrative extraction prompt.}
\label{tab:topicPrompt}
\renewcommand{\arraystretch}{1.3} 
\begin{tabular}{@{} p{0.20\linewidth} p{0.75\linewidth} @{}}
\hline
\raggedright \textbf{Component} & \textbf{Instruction} \\
\hline
\raggedright \textbf{Persona Assignment} & The model is instructed to be an ``AI threat intelligence and FIMI analyst'' to guide its reasoning into geopolitical manipulation. \\
\raggedright \textbf{Analytic Scope} & A strategic narrative was explicitly defined as a manipulative storyline that ``collapses uncertainty into a malicious plot'' (core difference to the task of extracting a topic). \\
\raggedright \textbf{Extraction Targets} & The model was required to identify three core elements from the cluster: the core claim, the enemy, and the specific manipulative angle (e.g., deliberate betrayal). \\
\raggedright \textbf{Few-Shot Calibration} & A complete example was provided, demonstrating how to synthesize raw cluster documents and keywords into structured reasoning and a final label. \\
\raggedright \textbf{Output Parameters} & A strict three-step execution leading to a one-sentence narrative label prefixed exactly with ``LABEL: ''. \\
\hline
\end{tabular}
\end{table}

\section{Evaluation}
\subsection{Validation and Boundary Analysis of Prompt-Based Filtering}
\begin{figure}
\includegraphics[width=.91\linewidth]{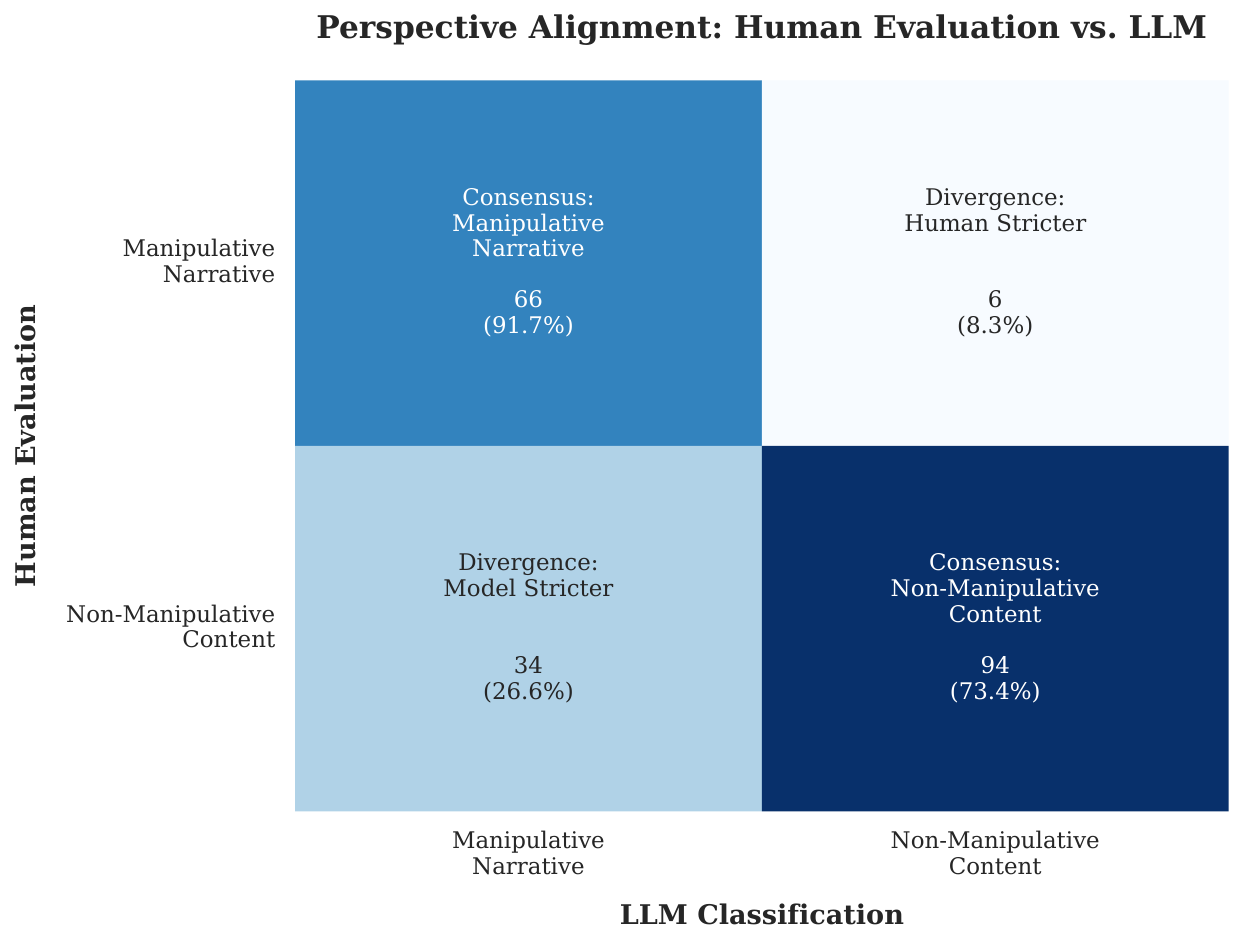}
\caption{Row-normalized alignment matrix, highlighting the model's high recall (91.7\%) and its tendency to be stricter than human raters.} \label{fig:confusion}
\end{figure}



To evaluate the reliability of the prompt-based filtering model, we conducted a two-stage manual audit on a balanced random sample of 200 posts (100 flagged by the model as manipulative narratives and 100 as non-manipulative content).

In the first stage, a human rater evaluated the dataset in a blind manner to mitigate confirmation bias. The posts were presented in random order, with the model's predictions hidden. The rater was tasked with determining whether each post contained fragments of a broader strategic narrative (e.g., ``social fragmentation'' or ``elite betrayal''), deliberately excluding legitimate political critique that lacked conspiratorial intent or allegations of a hidden agenda. To account for the inherent ambiguity of political discourse, the rater could classify highly ambiguous posts as ``borderline''. These borderline cases were systematically excluded, and replacement samples were drawn until the balanced 200-post corpus was fully restored.

In the second stage, a secondary evaluation of reasoning coherence was conducted. The rater was presented with the model’s final label alongside its generated reasoning to assess whether the model's logical deduction accurately aligned with its classification output.

The results of the classification audit are presented in Figure \ref{fig:confusion}. The prompt-based filtering achieved an F1 score of 0.77. Notably, the model exhibited a highly asymmetric performance profile, with a high recall of 0.92 but low precision of 0.66. As a result, the model prioritizes avoiding false negatives, accepting a higher rate of false positives to ensure that potential FIMI narrative fragments are not irretrievably discarded during the initial filtering stage. This high-recall bias is methodologically beneficial given the downstream pipeline architecture. Because HDBSCAN is a density-based algorithm that isolates outliers, falsely positive flagged posts are categorized as noise or fail to reach the semantic density required to form a cohesive strategic narrative cluster. Thus, the pipeline actively mitigates the impact of false positives originating from the filtering stage.

Regarding the secondary evaluation of reasoning coherence, the human rater agrees with the model's logical explanations in 95.5\% of cases. This high degree of agreement appears inconsistent with the F1 score of 0.77. However, this discrepancy highlights a fundamental challenge in FIMI detection: differentiating a coordinated manipulative narrative fragment from a genuine private yet highly populist opinion.

This boundary is best illustrated by examining a False Positive boundary case:

\textbf{Post (translated):} I’ll only vote for you if \#Remigration is included in your campaign platform.

\textbf{Reasoning (model): } Leverages the \#Remigration slogan to signal identity threat and demographic anxiety without addressing policy competence. This aligns with the ``Identity threat -> victimhood'' motif (Core Pillar 1), framing migration as an existential crisis to polarize the electorate and delegitimize mainstream parties that reject exclusionary narratives.

\textbf{Narrative Fragment (model): } True

\textbf{Narrative Fragment (user): } False

Based on the model's explanation, the reasoning is correct, since the demand for ``remigration'' aligns with the search for presumed identity threat and victimhood in the filtering prompt. However, from a human perspective, an individual citizen may hold the private political opinion that current migration levels are unsustainable, expressing this with a controversial slogan without being part of a coordinated external agenda. To conclude, the difference between human annotation and models' classification stems from the models' stringency, which makes it highly sensitive to polarization characteristics and tends to classify as true whenever a plausible justification exists, lacking the human mildness to dismiss such posts as private political expressions.

\subsection{Semantic Hyperparameter Tuning and Cluster Optimization}

One final hyperparameter to consider is the minimum cluster size for HDBSCAN, which specifies the minimum number of points required for a cluster to be recognized. In our HDBSCAN application, our goal is to reduce the number of posts classified as noise while ensuring that the assigned labels are sufficiently distinct. This means aiming for minimal noise-classified posts and maximizing the average distance between the narrative labels derived from each cluster. We evaluate several minimum cluster size options: 100, 200, 400, 600, 800, and 1000. The narratives from the clusters associated with each of these groupings are transferred into the embedding space using the Qwen3-Embedding-8B model as detailed in Section \ref{sec:embedding}, employing the same prompt as before: ``Identify the strategic narrative, manipulative intent, and underlying disinformation motive in the following text: ''. Furthermore, we calculate the ratio of posts labeled as noise to the total number of posts, which we aim to minimize. 
Table \ref{tab:hyperparametertuning} presents the outcomes of all tested configurations, indicating a sweet spot at a minimum cluster size of 400 posts. At this threshold, the noise level remains relatively low, while the average distance does not become excessively close. With a minimum cluster size of 400, the HDBSCAN algorithm successfully identified 41 distinct clusters.

\begin{table}[ht]
\centering
\setlength{\tabcolsep}{12pt} 

\caption{Hyperparameter Tuning Results for HDBSCAN. The optimal configuration (min\_cluster\_size = 400) minimizes data loss while maintaining a high average semantic distance (1 - cosine similarity) between extracted narratives.}
\vspace{2mm} 
\begin{tabular}{llll}
\hline
\textbf{Min. Size} & \textbf{Clusters} & \textbf{Noise (\%)} & \textbf{Avg. Semantic Distance} \\
\hline
100  & 118 & 43.80 & 0.3233 \\
200  & 75  & 40.04 & 0.3150 \\
\textbf{400}  & \textbf{41}  & \textbf{29.73} & \textbf{0.3057} \\
600  & 32  & 30.16 & 0.3054 \\
800  & 29  & 31.17 & 0.2953 \\
1000 & 24  & 32.19 & 0.2863 \\
\hline
\end{tabular}

\label{tab:hyperparametertuning}
\end{table}

\section{Results} \label{sec:results}
\begin{table}[!htbp]
\caption{The top 5 FIMI narrative clusters extracted by the pipeline, ranked by the volume of associated social media posts. The labels highlight the prevalence of deliberate betrayal and conspiracy motifs.}
\vspace{2mm}
\centering
\setlength{\tabcolsep}{6pt}
\begin{tabular}{r p{10.5cm}}
\hline
\textbf{Posts} & \textbf{Extracted Strategic Narrative} \\
\hline
15,379 & State Betrayal: The government deliberately endangers German children by permitting violent migrants entry and covering up the resulting crimes \\
10,450 & Deliberate Betrayal: The German government and Western elites are sacrificing national peace to fuel a proxy war against Russia for lobbyist profit \\
9,354  & Deliberate Betrayal: Corrupt health authorities and global elites are selling out the population to pharmaceutical corporations by forcing dangerous experimental vaccines \\
3,716  & Treacherous Elite Narrative: Mainstream media and political opponents are framed as treasonous servants of elite parties betraying the German people who will inevitably face violent revolutionary justice \\
3,266  & Western governments and media are complicit in Israeli genocide and white replacement because they are controlled by a malicious Zionist Jewish conspiracy \\
\hline
\end{tabular}
\label{tab:top5clusters}
\end{table}

Applying the optimized detection pipeline to the full social media corpus yielded 41 distinct narrative clusters surrounding German political figures. Table \ref{tab:top5clusters} presents the 5 largest clusters, their extracted strategic narratives, and their sizes. Since many of the listed strategic narratives share underlying motives, they can be manually grouped into four main thematic pillars.

\subsection{Pillar 1: The ``Great Replacement''}
This pillar aims to trigger existential fears by presenting migration as a weapon deliberately used by the government to destroy the native population and replace them with migrants.
Examples from our 41 automatically LLM-extracted clusters are:
\begin{itemize}
    \item State Betrayal: The government deliberately endangers German children by permitting violent migrants entry and covering up the resulting crimes.
    \item Civilizational Replacement: Complicit elites and Muslim immigrants are deliberately Islamizing Western nations to erase native culture and compromise security through resource diversion and institutional betrayal.
    \item Systemic Betrayal: Interior Minister Faeser is deliberately enabling an imported war by migrants while systematically persecuting German citizens through double standards and authoritarian overreach.
\end{itemize}

\subsection{Pillar 2: The Proxy War}
Germany is portrayed as a non-sovereign state controlled by the US. To undermine the actions against their own population, a zero-sum storyline is applied, arguing that supporting Ukraine would only destroy German prosperity.
\begin{itemize}
    \item Deliberate Betrayal: The German government and Western elites are sacrificing national peace to fuel a proxy war against Russia for lobbyist profit.
    \item Occupation Narrative: Germany remains a US colony since 1945, where collaborating political elites have sold out national sovereignty to Washington.
    \item German leadership and Western allies are deliberately sabotaging national prosperity and energy security by severing Russian gas ties to fund a proxy war in Ukraine.
\end{itemize}

\subsection{Pillar 3: The Climate Dictatorship}
Climate politics is presented as a fabricated hoax, invented by globalists to make citizens poorer and more controllable, to lead them into a totalitarian state.
\begin{itemize}
    \item The Green Party is portrayed as an illegitimate, anti-German sect imposing a climate dictatorship that deliberately harms the common people to serve elite ideological interests.
    \item Robert Habeck is framed as a hypocritical, communist-aligned antagonist deliberately destroying the German economy and threatening national sovereignty through incompetence and double standards.
    \item Deliberate Sabotage: The government and Green Party are intentionally destroying Germany's energy security and economic prosperity to enforce an ideological agenda against the common citizen.
\end{itemize}

\subsection{Pillar 4: The right-wing political party AfD as Savior}
The traditional parties are framed as a corrupt deep state that betrays its citizens, while the right-wing AfD and external actors like Elon Musk are the only remaining saviors.

\begin{itemize}
    \item Exclusive Salvation Narrative: The established parties and media enforce a fraudulent democracy to betray the citizens, positioning the AfD as the sole savior capable of curing the nation.
    \item Deep State Conspiracy: A globalist network controls all German parties except the AfD to deliberately destroy national sovereignty and civil liberties from within.
    \item External Savior Narrative: Elon Musk and global allies are backing the AfD to overthrow the corrupt German establishment and mainstream media.
\end{itemize}

\section{Discussion}

The results demonstrated that a classical topic modeling approach can be adapted to intent-driven strategic narrative clustering, in which the output is a storyline rather than a single word describing a topic. The adaptations made include mapping posts based on their manipulative intent, rather than their mere topic, into the embedding space by explicitly prompting the embedding model in that direction. This formed the basis for later narrative-based clustering, in which similar manipulative intents were grouped into the same cluster.

An important finding of our methodological evaluation is the synergy between the prompt-based filtering and the density-based clustering. The prompt-based reasoning classifier is highly stringent toward the provided FIMI strategies, yielding high recall (0.92) but lower precision (0.66) because any slight chance that one of the strategies is met causes the post to be flagged as positive. Without further processing, this approach, which produces many false positives, would be problematic. However, because the downstream HDBSCAN clustering algorithm can exclude noise, false positives will be filtered out, as they do not meet the characteristics of coordinated FIMI campaigns that form the large clusters.

Another finding is the benefit of large reasoning models to incorporate their trained knowledge into the provided few-shot classification tasks. In contrast, traditional supervised classifiers are limited to the vocabulary and features presented in their training data. Our analysis of the filtering phase uncovered that the LLM actively applied its own Open Source Intelligence (OSINT) knowledge to evaluate social media posts. For example, the model correctly recognized the domain ``rtde.media'' as a proxy for the Russian state-controlled outlet ``Russia Today (RT)'', resulting in the flagging of the underlying post. Since this domain was not mentioned in the filtering prompt, this demonstrates that the model connected the filtering task with its own world knowledge.

Although these results are promising, the illustrated approach also has its limitations. The most significant might be the blurry line between coordinated, manipulative, strategic narratives and populist, radical, yet personal political views. Additionally, the shown filtering prompt is the result of an iterative human-AI collaboration. While effective, this manual prompt engineering still relies on human intuition regarding campaigns, characteristics, and which few-shot examples should be included in the filtering prompt to best guide the model.

\section{Outlook}

To address the limitations of manual prompt design, future work might focus on transitioning from heuristic prompt engineering to programmatic prompt optimization. A foundation might be an annotated ground-truth corpus comprising posts on FIMI and non-FIMI content.
Based on that corpus, optimization frameworks can systematically search for optimal prompts and few-shot examples to achieve higher precision and recall.

Future research could delve deeper into the origins and development of specific narratives. This may involve a focused examination of clustering outputs across web and social media platforms to gain insights into the dynamics of these campaigns. Such analysis can also aid in identifying the sources of manipulative content and understanding which individuals are particularly vulnerable to it.

Ultimately, FIMI should be considered within a broader framework that includes the use of visual language models (VLMs) and the assessment of posts against a repository of reliable sources. This approach would enable the processing of memes, manipulated images, and videos, along with the contextual knowledge needed to evaluate the authenticity of news, despite their seemingly obscure nature. Such a strategy could serve as a robust defense mechanism against the next wave of disinformation and malinformation warfare.

\section{Data and Code Availability}
To support scientific transparency and reproducibility, the source code for the FIMI detection pipeline, is available at: \newline \url{https://github.com/SinclairSchneider/manipulative\_narrative\_detection}.

\section{Ethical Considerations and Statement of Objectivity}
Researching Foreign Information Manipulation and Interference (FIMI) involves analyzing polarizing content. The strategic narratives in this paper, particularly in Section \ref{sec:results} and Table \ref{tab:top5clusters}, are unedited outputs from the LLM pipeline for transparency and reproducibility. These narratives do not reflect the authors' views, who reject any political opinions, conspiracy theories, or manipulative claims in the dataset.

\begin{credits}
\subsubsection{\ackname} The authors would like to thank the System Sciences Chair for Communication Systems and Network Security under the direction of Prof. Dr. Gabi Dreo Rodosek.



\subsubsection{\discintname}
The authors have no competing interests to declare that are
relevant to the content of this article.
\end{credits}

\bibliographystyle{splncs04}
\bibliography{mybibliography}

\end{document}